\documentclass[manuscript,screen]{acmart}
\usepackage{verbatim}
\usepackage{multirow}
\usepackage{caption}
\usepackage{graphicx}
\usepackage{ulem}
\usepackage{balance}
\usepackage[misc]{ifsym}
\usepackage{subfigure} 

\usepackage{amssymb}

\usepackage{amsmath, bm}
\usepackage{hyperref}
\hypersetup{
    colorlinks=true,
    linkcolor=blue,
    filecolor=blue,      
    urlcolor=blue,
    citecolor=black,
}

\usepackage{epstopdf}
\usepackage{makecell}

\AtBeginDocument{%
  \providecommand\BibTeX{{%
    \normalfont B\kern-0.5em{\scshape i\kern-0.25em b}\kern-0.8em\TeX}}}

\setcopyright{acmcopyright}
\copyrightyear{2021}
\acmYear{2021}
\acmDOI{10.1145/1122445.1122456}

\acmConference[AISS '21]{AISS '21: 3rd International Conference on Advanced Information Science and System}{November 26--28, 2021}{AISS, NY}
\acmBooktitle{AISS '21: 3rd International Conference on Advanced Information Science and System,
  November 26--28, 2021, AISS, NY}
\acmPrice{15.00}
\acmISBN{978-1-4503-XXXX-X/18/06}

\begin{document}

\title{HENet: Forcing a Network to Think More for Font Recognition}

\author{Jingchao Chen}
\email{ArthurN@shu.edu.cn}
\orcid{0000-0002-1617-7930}
\affiliation{%
  \institution{Shanghai University}
  \streetaddress{P.O. Box 1212}
  \city{Shanghai}
  \state{Shanghai}
  \country{China}
  \postcode{200444}
}

\author{Shiyi Mu}
\affiliation{%
  \institution{Shanghai University}
  \streetaddress{P.O. Box 1212}
  \city{Shanghai}
  \state{Shanghai}
  \country{China}
  \postcode{200444}
}

\author{Shugong Xu}
\email{shugong@shu.edu.cn}
\affiliation{%
  \institution{Shanghai University}
  \streetaddress{P.O. Box 1212}
  \city{Shanghai}
  \state{Shanghai}
  \country{China}
  \postcode{200444}
}

\author{Youdong Ding}
\email{ydding@shu.edu.cn}
\affiliation{%
  \institution{Shanghai University}
  \streetaddress{P.O. Box 1212}
  \city{Shanghai}
  \state{Shanghai}
  \country{China}
  \postcode{200444}
}

\renewcommand{\shortauthors}{Chen, et al.}

\begin{abstract}
Although lots of progress were made in Text Recognition
/OCR in recent years, the task of font recognition is remaining challenging.  The main challenge lies in the subtle difference between these similar fonts, which is hard to distinguish. This paper proposes a novel font recognizer with a pluggable module solving the font recognition task. The pluggable module hides the most discriminative accessible features and forces the network to consider other complicated features to solve the hard examples of similar fonts, called HE Block. Compared with the available public font recognition systems, our proposed method does not require any interactions at the inference stage. Extensive experiments demonstrate that HENet achieves encouraging performance, including on character-level dataset Explor all and word-level dataset AdobeVFR.
\end{abstract}



\maketitle

\section{Introduction}
As is known to us, different contents in the scanned document are always written in various fonts. Thus, font information in scanned document images contains rich semantic information among texts that is very useful for understanding essential information of the whole document image, such as bank application forms, receipts, and insurance claim forms. In addition, font is essential for many media workers. They often need to create a poster or write an article with attractive fonts. The best choice to find a suitable font for these media workers is to take a photo of the desired text style and seek help on font-recognizing websites. Most font websites demand complicated interactions for users and a series of pre-processing steps on images, but the recognition results are still inferior. The proposed method provides a solution without tedious pre-processing processes and interactions for font images, which still achieves a satisfactory performance.

With the rapid development of deep learning algorithms, great success has been achieved on many computer vision tasks(e.g., image classification, object detection, and optical character recognition). Font recognition can be regarded as a special task of image classification, which takes a raw image as input, and subsequently learns its class-specific feature representation through a CNN-based network. Finally, the result can be predicted with the class-specific feature representation through the classifier layer. The current research on the use of CNNs for deep learning in computer vision has got many encouraging results, including AlexNet \cite{krizhevsky_imagenet_2017} , ResNet \cite{he2016deep} , VGGNet \cite{simonyan_very_2015} , GoogleNet \cite{Szegedy_2015_CVPR} and others. However, directly using these networks on font recognition definitely can not get a satisfactory performance. Some task-specific methods are needed for the task of font recognition.

\begin{figure}[ht]
\centering
\includegraphics[width=230pt]{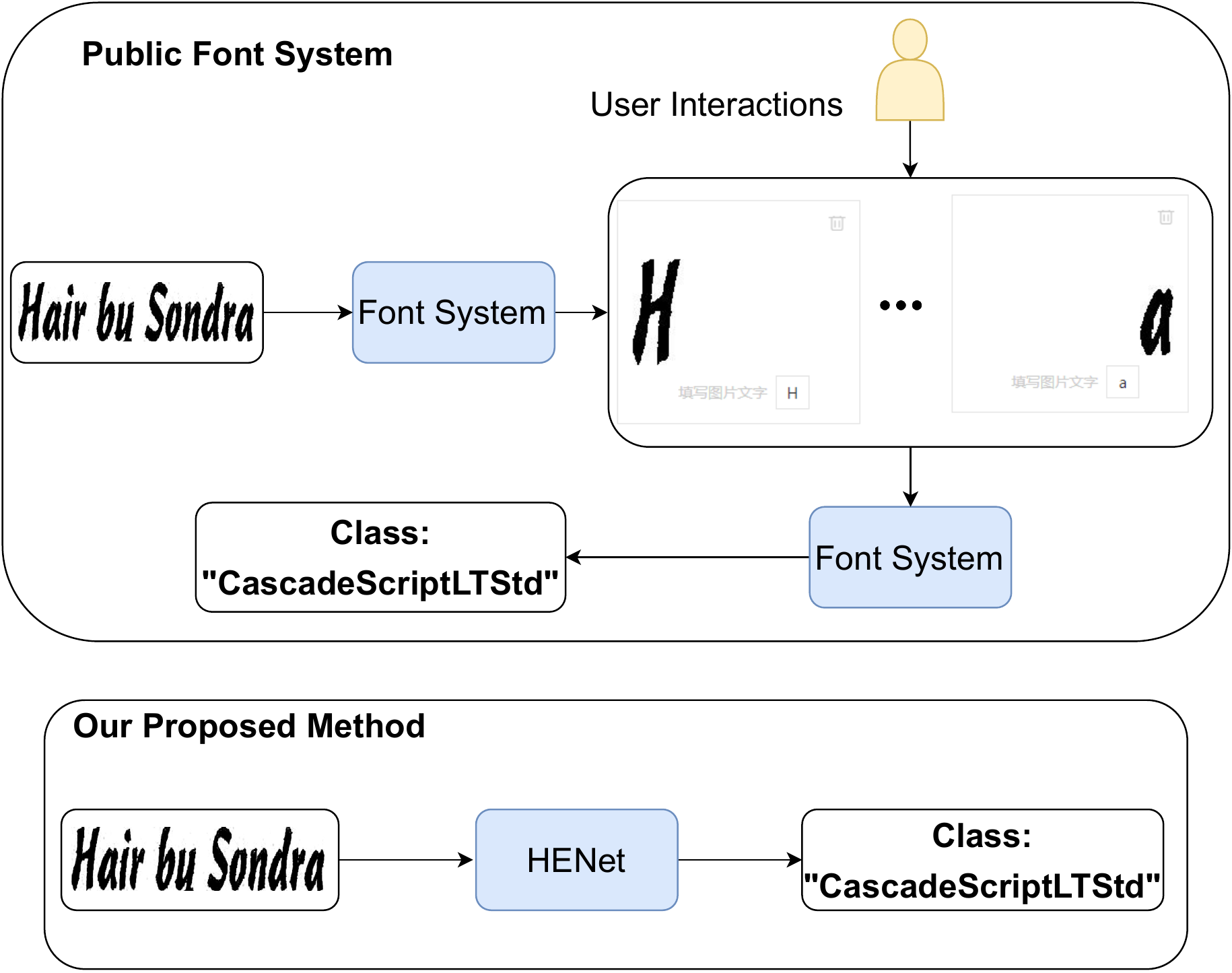}
\caption{Comparison between the pipeline of public font website and our proposed method. Most public font website requires users to enter each character for segmentation, while our pipeline does not need all these processes which is an end to end system.}
\label{fig1}
\end{figure}

In contrast to the general image classification, font recognition is more challenging due to the larger space of candidate fonts(i.e., Available datasets usually contain thousands of font classes). Furthermore, texts in different fonts always have a similar overall appearance, only varying in specific stroke style details of individual characters. These differences in fonts can not even be distinguished for an ordinary person. As the popular deep learning-based methods generally are data-driven, the collection of labeled fonts proves to be very difficult. What is more, an ordinary person is not qualified for the labeling task on error-prone font images. All of the above difficulties require the network to explore complicated features under accessible features on limited input font images.  

Inspired by the great success of deep learning models in various computer vision tasks (e.g., image classification). We proposed a font recognition method for Latin alphabets based on the Convolutional Neural Network called HENet.  As shown in Fig.~\ref{fig1}, HENet does not depend on any information about character segmentation or character recognition and obtains a significant performance improvement on Explor\_all Dataset covering thousands of font classes. To summarize, our contributions are as follows:

\begin{itemize}
\item We present a new approach for font recognition, which is called HENet. The network is end to end and does not depend on any extra input information apart from the text image, which is convenient for people to use to a large extent.
\item We propose a pluggable module named HE Block to improve the accuracy of font recognition for HENet.The HE Block suppresses features with the most prominent response values and compels the network to find more complicated features to make a correct prediction on similar fonts. 
\item We conduct several experiments, which demonstrate that HENet achieves high accuracy on the character-level dataset Explor\_all and word-level dataset Adobe\_VFR.
\item We further validate the effectiveness and generalization on different backbone networks and different datasets through some experiments.
\end{itemize}


\section{Related Work}
\subsection{Font Recognition}
Font recognition has been explored in the past few years as part of the document analysis system. Therefore many online font recognizers that require users to upload a font image and enter characters one by one for algorithms like pattern matching, have been released with these researches on font recognition, as shown in Fig.~\ref{fig1}. These font interfaces, such as rightknights and likefont, need user interactions and relatively complicated image processing steps. 

Researchers generally regard the visual font recognition task as a special image classification task with thousands of similar classes to be distinguished. Carlos et al. \cite{aviles-cruz_high-order_2005} proposed global texture analysis on font recognition that employs the sliding window analysis method to obtain the features of the document, using fourth and third-order moments. Tao et al. \cite{tao_sparse_2014} applied LBP descriptor-based Chinese character interesting points for representing discriminative font information to build a Chinese font recognition system. The above studies contain two stages(feature extraction and classifier design) that are not end-to-end and need prior knowledge about the data domain.

Recently, deep learning-based approaches have been emphasized significantly in the fields of computer vision. The researchers who study on font recognition change their attention from the traditional two-stage method to new CNN-based high-performance algorithms. To the best of our knowledge, there are researches made on font recognition from different aspects. Wang et al. \cite{wang_deepfont_2015} proposed such a method that utilizes a Convolutional Neural Network and a domain adaptation technique based on SCAE. The method requires a large amount of unlabeled real-world data and millions of synthetic data, which results in a colossal training cost. \cite{wang_deepfont_2015} have also collected a font dataset named AdobeVFR. We also evaluate our method on this word-level font dataset. Huang et al. \cite{huang_dropregion_2018} propose a font recognizer for Chinese characters and Chinese text blocks, which is made up of a modified inception module and convolutions. Due to the lack of diversity of the font data, the method of dropregion was proposed to generate a large number of stochastic variant font samples whose local regions are selectively disrupted. Zhang et.al\cite{DBLP:journals/ijdar/ZhangGF19}proposed a method on Chinese calligraphy styles which combines Squeeze-and-excitation block and Haar transform layer in Convolution networks. Yang et.al\cite{DBLP:journals/ijdar/YangKKK19}proposed a Hangul font cluster recognizer to address the issues caused by indistinguishable fonts and untrained new fonts. Goel N et al. \cite{Goel_2020_CVPR_Workshops} combines the recent few-shot learning technique like prototype network with font recognition to achieve the effect of identifying new fonts with very few annotated samples per font. Our proposed HENet utilizes a novel module to regularize the feature extraction network, partially restraining the most prominent feature response in class activation maps to explore complicated features for hard examples. Therefore, the subtle difference in similar fonts can be found and is utilized to make a correct prediction.

\begin{figure}[ht]
\centering
\includegraphics[width=270pt]{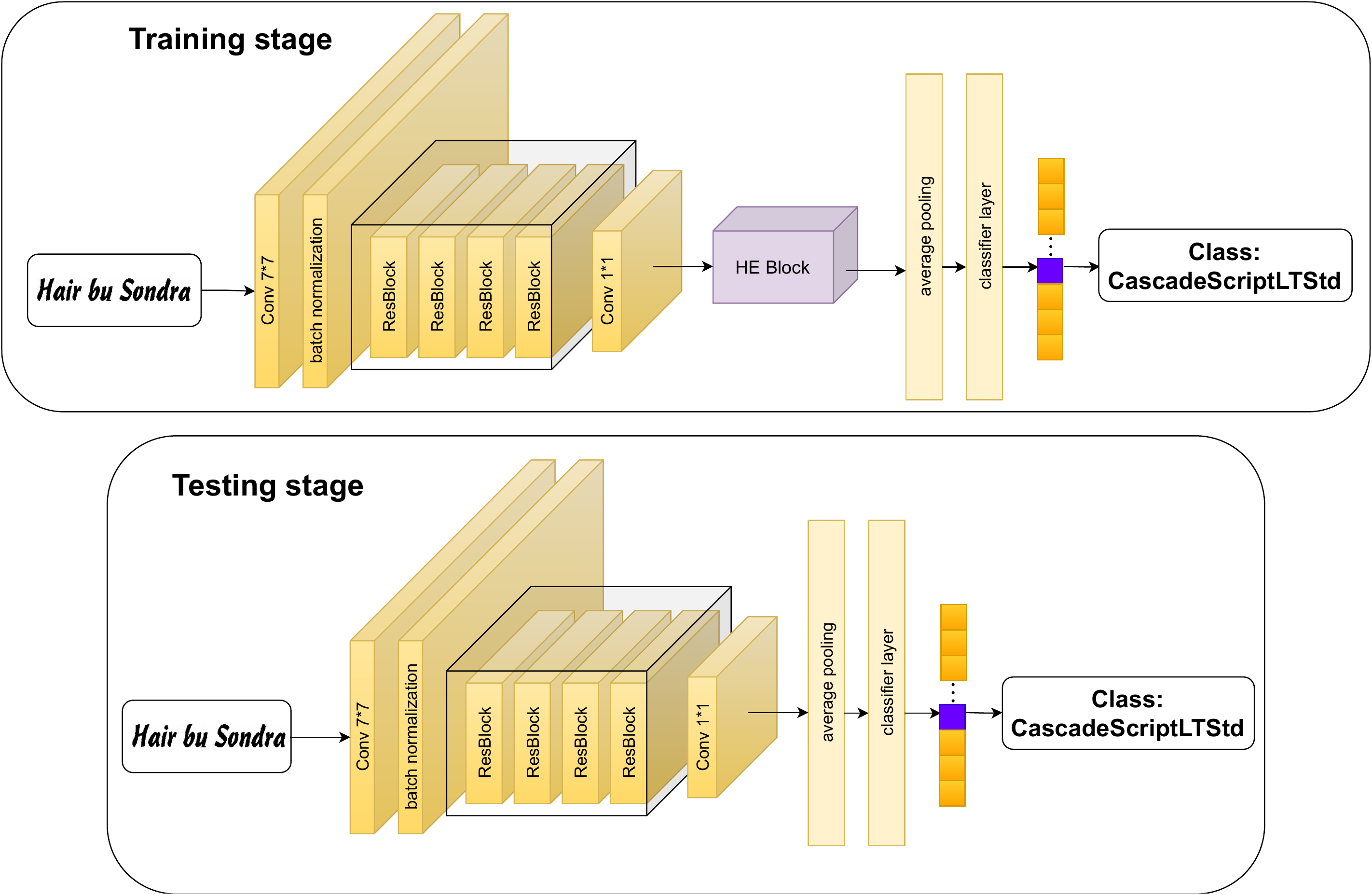}
\caption{Overview of our overall architecture. Our pipeline contains three components in the training phase: a backbone for feature extraction, HE Block, and a classifier. The HE Block suppresses the most discriminative regions of the category-specific activation maps and forces the network to think more and find alternative complicated features. As a result, the network achieves the best performance by differentiating the similar font classes with more informative regions. The font class can be predicted without HE Block in the testing phase, which does not add extra computation cost.}
\label{fig2}
\end{figure}

\section{Method}
In this section, we introduce our proposed font recognizing method, which aims to solve the indistinguishable nature of those similar font classes. The architecture of the proposed HENet, as shown in Fig.~\ref{fig2}, contains three components, the backbone of feature extraction, HE Block, and font classifier. The feature extraction network learns a preliminary feature representation from the raw font image. These extracted features are sent to HE Block to suppress some of the most discriminative features, forcing the backbone of feature extraction to think more and find other informative features of strokes and text styles. Eventually, the font classifier can predict a more accurate result by mining the information hidden under the most discriminative feature.

\subsection{HE Block}
As the task of font recognition is a multi-classification task with C classes as shown in Fig.~\ref{fig2}. (x, y) denotes the image-label pair from the dataset (X, Y), where X and Y are the image dataset and the label collection of all C classes. The input of the HE Block $F \in R^{C \times H \times W}$ is the class-specific activation map that is extracted from the previous network. It needs to be pointed out that F = \{$F_c$: $c \in $[1, C]\}, where $F_c \in R^{H \times W}$ is an individual activation map which represents the  $c^{th}$ class. H and W denote the size(height and width) of the output from the feature extraction network. 

The concept of HE Block means "Hide and Enhance". In the training phase, we utilize the HE Block to hide a part of the highest response of the category-specific activation map, that is to say, suppress the accessible feature(the alignment and style of the global text). Then the feature extraction is obliged to learn other helpful subtle differences(the stroke details) to make the correct prediction, which helps enhance the performance of font recognition.

\begin{figure}[ht]
\centering
\includegraphics[width=250pt]{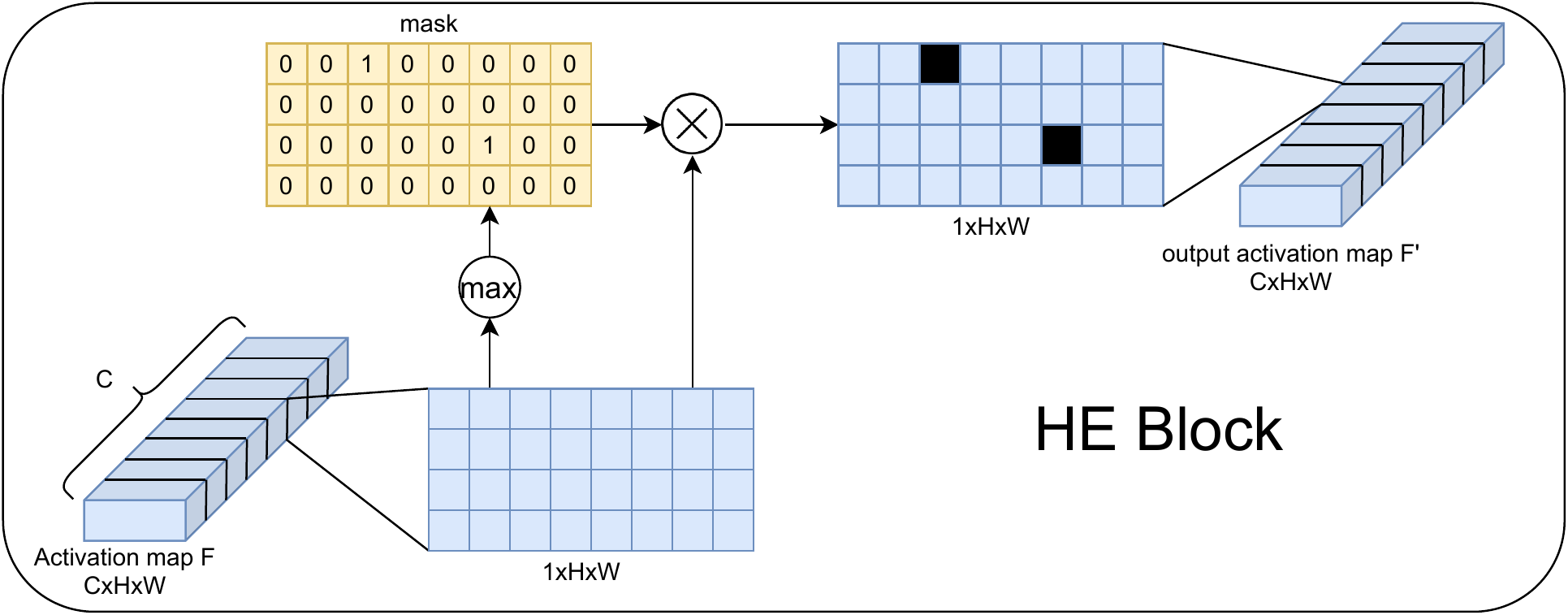}
\caption{the details of the HE Block}
\label{fig3}
\end{figure}

\subsubsection{Mask Creation} The process of creating a mask for the category-specific activation map is introduced in this section. For an input activation map $F \in R^{C \times H \times W}$, $M \in R^{C \times H \times W}$ denotes the mask whose values are binary, representing whether the corresponding location in F needs to be hidden. For every element M\_c in M, 1 means that the value of the corresponding location in M\_c will be hidden, while 0 means the activation at this place will not change completely.

\paragraph{Maximum Suppression} First, we randomly suppress the maximum responses in the category-specific activation map. These features mainly pay attention to the global texture information that is easy to learn during the current training period. After these global texture features are suppressed, the ability of the whole recognizer correspondingly decreases a bit in this iteration. However, the regularization of the cross-entropy loss forced the network to think more and look for other complicated information like subtle stroke features from other regions, making the network explore the semantics of the input image further. Let $P_c \in R^{H \times W}$ be the maximum mask, corresponding to the $c^{th}$ class of the activation map F, can be expressed by formula as follows:

\begin{equation}
M_c(i, j) = \left\{
\begin{aligned}
& 1 \qquad {\rm if \ F_c(i, j) = max(F_c)}\\
& 0 \qquad {\rm otherwise}\\
\end{aligned}
\right
.
\end{equation}

Here, max($F_c$) means the maximum response of the category-specific activation map $F_c$. We randomly hide the max values of $c^{th}$ channel in the whole activation map F.

\subsubsection{Weight of Mask} Suppose we set the weight of the mask $M_c$ as 0. In that case, it will not achieve excellent performance since the complete loss of the most prominent discriminate feature necessarily leads to a significant drop in performance. Let $F^{'}$ be the output of the HE Block, which is denoted as follows:

\begin{equation}
F_c^{'}(i, j) = \left\{
\begin{aligned}
&  F_c(i, j) \quad & {\rm if \ M_c(i, j) = 0}\\
& \beta * F_c(i, j) \quad & {\rm if \ M_c(i, j) = 1}\\
\end{aligned}
\right
.
\end{equation}

where $\beta$ means the weight of the initial activation map $F_c$. In other words, as the input of the final classifier, the feature map is substituted with their initial values by the weight $\beta$ when the mask of the corresponding location has a value of 0. In the section of experiment, we set the weight as 0.5, which gives the best performance.

\subsection{Training Strategy for Font Recognition}
It is worth mentioning that we need to replace the global average pooling and fully connected layer in the origin backbone network with $1\times1$ convolution to apply our method, whose output channel is the same as the number of classes. Correspondingly, the output of our backbone network for feature extraction has a size of $C \times H \times W$, instead of $C \times 1 \times 1$. After the process of HE Block, we resend the feature map to an average pooling layer and get confidence score of prediction on the font image, which is shown as follows:

\begin{equation}
S_c = Avgpool(F_c^{'})
\end{equation}

Where $S_c$ means the confidence score for each class. Here, AvgPool denotes the operation of average pooling layer.

Our approach is an end-to-end font recognizer. In the training stage, the activation maps are sent to our proposed HE Block. Then we get the confidence score for each font class through the global average pooling layer, which is used to make a prediction. The accessible global feature is hidden by HE block. Thus the network learns other class-sensitive complicated features covered up by accessible features, which effectively enhances the performance of the recognizer. While in the testing phase, HE Block is discarded so that the whole activation map can directly pass through the global average pooling. And then, all of the exploited discriminative features learned by the previous network in the training phase make their effort together to make the final prediction.

\section{Experiment}
\subsection{Datasets}
Font recognition is a challenging task due to the subtle difference between the similar candidate font classes, which increases recognition difficulty. We adopt the font dataset explor\_all released by O’Donovan et al. \cite{odonovan_exploratory_2014} which consists of 69192 images belonging to 1116 font classes to evaluate our algorithm on character-level font recognition. In the dataset Explor\_all, each font class contains 62 images(10 numbers and 52 Latin alphabets). In our experiment, we split the whole dataset for training and testing. We randomly choose 10 images in each font class for testing and treat the rest as a training set. The task becomes more challenging because the network has not seen characters divided into the test set in the training period. Thus the network has to learn the class-specific feature for each font class, which requires the network to explore the feature space of font classes further and has the ability to generalize.

What's more, we also evaluate our method on the word-level font dataset AdobeVFR, which was released by Wang et al. \cite{wang_deepfont_2015}. The dataset of AdobeVFR comprises 2383 font classes, and each font among them contains 1000 synthetic images. The content of word in these images was sampled from a large corpus. Compared to the character-level font dataset, the word-level dataset has more features to be extracted for font recognition. However, the real-world test set has an extremely great difficulty due to the domain shift. We use AdobeVFR to evaluate the font recognizing performance on word images.

\subsection{Implementation Details}
We use the resolution of 64$\times$64 for our character-level font recognition experiments, while the resolution of 64$\times$256 for our word-level font recognition experiments due to the special ratio of the word images. We trained our network whose backbone is modified Resnet18 \cite{he2016deep}. Momentum SGD optimizer is utilized with an initial learning rate of 0.001, which decays by 0.8 for every 5 epochs. We set weight decay as $10^{-4}$. Our algorithm is implemented using Pytorch \cite{paszke2017automatic} with a GeForce RTX 2080 Ti.

\subsection{Quantitative Results}
Our method does not require any extra parameters and helps the network find those subtle stroke features that are difficult to learn. Our results are compared with the most recent font recognition approaches on the Explor\_all Dataset with a similar experimental setting. We also compared our method with dropout because they both improve performance by masking some features. Dropout randomly masks some neurons for fully connected layers, while our HE block masks the maximum response and partially preserves information preservation. The comparisons results with these state-of-the-art methods are shown in Table~\ref{tab1}

We observe that our method achieves the best accuracy on Explor\_all Dataset. We achieve 86.31\% top-1 accuracy surpassing DropRegion\cite{huang_dropregion_2018}(85.71\%). DropRegion\cite{huang_dropregion_2018} obtain a good performance because they modify the inception module by adding 3 branches and uses the augmentation method dropregion. Our method weakens accessible features like texture and overall appearance, forcing the network to learn more complicated knowledge like subtle stroke differences. It helps the recognizer perform better on distinguishing similar font classes.

\subsection{Ablation Study}
To comprehensively analyze our method, we conduct related experiments to verify the setting of the parameters. We conduct our ablation studies on Explor\_all dataset.

Here, we first confirm the effectiveness of our proposed HE Block on different backbones, and all of them achieve a significant performance improvement. As shown in Table~\ref{tab2}, we can see that in the experiments of different backbone networks, HE Block improves top-1 accuracy by 0.86\% for ResNet18, 0.44\% for ResNet34, 0.56\% for ResNet50, 0.44\% for ResNet101.

Then, we show our ablation study analysis on the weight of mask $\beta$. Top-1 and Top-5 accuracy for different weight $\beta$ settings is shown in Table~\ref{tab3}. It demonstrates that setting $\beta$ as 0.5 leads to the best performance among experiments and outperforming the network without HE Block( $\beta$ = 1 ) by 0.86\%. For top-5 accuracy, setting the weight to 0.7 gives the best performance, which also makes an improvement of 0.44\%.  
\begin{table}
\caption{Experimental results on Explor\_all between our network and other state-of-the-art font recognition models are shown. Our method outperforms existing approaches recognition accuracy.}\label{tab1}
\centering
\begin{tabular}{|c|c|c|}
\hline
\makecell[c]{Methods} & \makecell[c]{Accuracy(top-1)} & \makecell[c]{Accuracy(top-5)}  \\
\hline
\makecell[c]{CalliNet\cite{DBLP:journals/ijdar/ZhangGF19}} & \makecell[c]{65.37}  & \makecell[c]{92.50}\\
\hline
\makecell[c]{Dropout\cite{DBLP:journals/jmlr/SrivastavaHKSS14}} &  \makecell[c]{72.33}  & \makecell[c]{94.87}\\
\hline
\makecell[c]{Hanfont\cite{DBLP:journals/ijdar/YangKKK19}} &  \makecell[c]{76.86}   & \makecell[c]{95.88}\\
\hline
\makecell[c]{DropRegion\cite{huang_dropregion_2018}} &  \makecell[c]{85.71}  &  \makecell[c]{98.29} \\
\hline
\makecell[c]{ours}  &  \makecell[c]{{\bfseries 86.31 }}  & \makecell[c]{98.48}\\
\hline
\end{tabular}
\end{table}

\begin{table}
\caption{HE Block performance with different backbones}\label{tab2}
\centering
\begin{tabular}{|c|c|c|c|c|c|c|c|c|}
\hline
\makecell[c]{Backbone}&
\multicolumn{2}{c|}{ResNet18}&
\multicolumn{2}{c|}{ResNet34}&
\multicolumn{2}{c|}{ResNet50}&
\multicolumn{2}{c|}{ResNet101}\\
\hline
\makecell[c]{Accuracy}  & top-1 &top-5 &top-1 &top-5 & top-1 &top-5 & top-1 &top-5 \\
\hline
w/o HE Block & 85.45 & 98.24 & 82.87 & 97.56 & 78.70& 96.25 &78.73 & 96.01\\
\hline
w/ HE Block & {\bfseries 86.31} & {\bfseries 98.68} & {\bfseries 83.31} & {\bfseries 97.60} & {\bfseries 79.26}& {\bfseries 96.74} & {\bfseries 79.17}& {\bfseries 96.61}\\
\hline
\end{tabular}
\end{table}

\begin{table}
\caption{Ablation study on weight of mask. Setting weight to 0.5 makes the best performance.}\label{tab3}
\centering
\begin{tabular}{|c|c|c|}
\hline
\makecell[c]{Weight of Mask} & \makecell[c]{Accuracy(top-1)} & \makecell[c]{Accuracy(top-5)}  \\
\hline
\makecell[c]{1} & \makecell[c]{85.45}  & \makecell[c]{98.24}\\
\hline
\makecell[c]{0.9} &  \makecell[c]{84.61}  & \makecell[c]{98.21}\\
\hline
\makecell[c]{0.8} &  \makecell[c]{85.44}  & \makecell[c]{98.43}\\
\hline
\makecell[c]{0.7} &  \makecell[c]{85.63}  & \makecell[c]{\bfseries 98.68}\\
\hline
\makecell[c]{0.6} &  \makecell[c]{85.91}   &  \makecell[c]{98.43} \\
\hline
\makecell[c]{0.5}  &  \makecell[c]{{\bfseries 86.31 }}  & \makecell[c]{98.48}\\
\hline
\makecell[c]{0.4} &  \makecell[c]{86.18}   & \makecell[c]{98.43}\\
\hline
\makecell[c]{0.3} &  \makecell[c]{85.07}  &  \makecell[c]{98.38} \\
\hline
\end{tabular}
\end{table}

Finally, we also evaluate our algorithm on the AdobeVFR dataset, which is a word-level font dataset. In Table~\ref{tab4}, the experiment demonstrates that HE Block improves top-1 performance on synthetic validation set as well as the real-world test set by 0.61\% and 1.14\%. Although The real-world test set is extraordinarily challenging and the training set contains no real-world data, our method still brings considerable improvement on word-level.

\begin{table}
\caption{Evaluation on AdobeVFR dataset.}\label{tab4}
\centering
\begin{tabular}{|c|c|c|c|c|}
\hline
\makecell[c]{Accuracy} & \makecell[c]{syn(top-1)} & \makecell[c]{syn(top-5)} & \makecell[c]{real(top-1)} & \makecell[c]{real(top-5)}\\
\hline
\makecell[c]{ResNet18}   & \makecell[c]{97.62}  & \makecell[c]{99.95}  & \makecell[c]{46.27}  & \makecell[c]{63.35}\\
\hline
\makecell[c]{ResNet18+HE Block} & \makecell[c]{{\bfseries 98.23}} &  \makecell[c]{{\bfseries 99.98}} &  \makecell[c]{{\bfseries 47.41}} &  \makecell[c]{{\bfseries 65.11}}\\
\hline
\end{tabular}
\end{table}

\section{Conclusion}
In our paper, we proposed a method to get better performance on font recognition whose categories are very similar. Our proposed HE Block suppresses the prominently accessible feature so that the whole network can think more and learn more complicated stroke details. What's more, the utilization of HE Block is pluggable during the training phase and does not bring any extra computation cost during inference time. We conduct related experiments with different backbone networks, proving that our method works on different backbones and surpasses the state-of-the-art font recognition models. In our future work, we will make a large-scale real-world font dataset containing both Chinese and Latin characters as a public benchmark for font recognition. Then, we will further explore the popular transformer-based model to obtain global and local strokes information.

\bibliographystyle{ACM-Reference-Format}
\bibliography{HENet}


\begin{thebibliography}{14}


\ifx \showCODEN    \undefined \def \showCODEN     #1{\unskip}     \fi
\ifx \showDOI      \undefined \def \showDOI       #1{#1}\fi
\ifx \showISBNx    \undefined \def \showISBNx     #1{\unskip}     \fi
\ifx \showISBNxiii \undefined \def \showISBNxiii  #1{\unskip}     \fi
\ifx \showISSN     \undefined \def \showISSN      #1{\unskip}     \fi
\ifx \showLCCN     \undefined \def \showLCCN      #1{\unskip}     \fi
\ifx \shownote     \undefined \def \shownote      #1{#1}          \fi
\ifx \showarticletitle \undefined \def \showarticletitle #1{#1}   \fi
\ifx \showURL      \undefined \def \showURL       {\relax}        \fi
\providecommand\bibfield[2]{#2}
\providecommand\bibinfo[2]{#2}
\providecommand\natexlab[1]{#1}
\providecommand\showeprint[2][]{arXiv:#2}

\bibitem[\protect\citeauthoryear{Avilés-Cruz, Rangel-Kuoppa, Reyes-Ayala,
  Andrade-Gonzalez, and Escarela-Perez}{Avilés-Cruz et~al\mbox{.}}{2005}]%
        {aviles-cruz_high-order_2005}
\bibfield{author}{\bibinfo{person}{Carlos Avilés-Cruz}, \bibinfo{person}{Risto
  Rangel-Kuoppa}, \bibinfo{person}{Mario Reyes-Ayala}, \bibinfo{person}{A.
  Andrade-Gonzalez}, {and} \bibinfo{person}{Rafael Escarela-Perez}.}
  \bibinfo{year}{2005}\natexlab{}.
\newblock \showarticletitle{High-order statistical texture analysis––font
  recognition applied}.
\newblock \bibinfo{journal}{\emph{Pattern Recognition Letters}}
  (\bibinfo{date}{Jan.} \bibinfo{year}{2005}).
\newblock


\bibitem[\protect\citeauthoryear{Goel, Sharma, and Vig}{Goel
  et~al\mbox{.}}{2020}]%
        {Goel_2020_CVPR_Workshops}
\bibfield{author}{\bibinfo{person}{Nikita Goel}, \bibinfo{person}{Monika
  Sharma}, {and} \bibinfo{person}{Lovekesh Vig}.}
  \bibinfo{year}{2020}\natexlab{}.
\newblock \showarticletitle{Font-ProtoNet: Prototypical Network-Based Font
  Identification of Document Images in Low Data Regime}. In
  \bibinfo{booktitle}{\emph{Proceedings of the IEEE/CVF Conference on Computer
  Vision and Pattern Recognition (CVPR) Workshops}}.
\newblock


\bibitem[\protect\citeauthoryear{He, Zhang, Ren, and Sun}{He
  et~al\mbox{.}}{2016}]%
        {he2016deep}
\bibfield{author}{\bibinfo{person}{Kaiming He}, \bibinfo{person}{Xiangyu
  Zhang}, \bibinfo{person}{Shaoqing Ren}, {and} \bibinfo{person}{Jian Sun}.}
  \bibinfo{year}{2016}\natexlab{}.
\newblock \showarticletitle{Deep residual learning for image recognition}. In
  \bibinfo{booktitle}{\emph{CVPR}}.
\newblock


\bibitem[\protect\citeauthoryear{Huang, Zhong, Jin, Zhang, and Wang}{Huang
  et~al\mbox{.}}{2018}]%
        {huang_dropregion_2018}
\bibfield{author}{\bibinfo{person}{Shuangping Huang}, \bibinfo{person}{Zhuoyao
  Zhong}, \bibinfo{person}{Lianwen Jin}, \bibinfo{person}{Shuye Zhang}, {and}
  \bibinfo{person}{Haobin Wang}.} \bibinfo{year}{2018}\natexlab{}.
\newblock \showarticletitle{{DropRegion} training of inception font network for
  high-performance {Chinese} font recognition}.
\newblock \bibinfo{journal}{\emph{Pattern Recognition}} (\bibinfo{date}{May}
  \bibinfo{year}{2018}).
\newblock


\bibitem[\protect\citeauthoryear{Krizhevsky, Sutskever, and Hinton}{Krizhevsky
  et~al\mbox{.}}{2017}]%
        {krizhevsky_imagenet_2017}
\bibfield{author}{\bibinfo{person}{Alex Krizhevsky}, \bibinfo{person}{Ilya
  Sutskever}, {and} \bibinfo{person}{Geoffrey~E. Hinton}.}
  \bibinfo{year}{2017}\natexlab{}.
\newblock \showarticletitle{ImageNet classification with deep convolutional
  neural networks}.
\newblock \bibinfo{journal}{\emph{Commun. ACM}} (\bibinfo{date}{May}
  \bibinfo{year}{2017}), \bibinfo{pages}{84--90}.
\newblock


\bibitem[\protect\citeauthoryear{O'Donovan, Lībeks, Agarwala, and
  Hertzmann}{O'Donovan et~al\mbox{.}}{2014}]%
        {odonovan_exploratory_2014}
\bibfield{author}{\bibinfo{person}{Peter O'Donovan}, \bibinfo{person}{Jānis
  Lībeks}, \bibinfo{person}{Aseem Agarwala}, {and} \bibinfo{person}{Aaron
  Hertzmann}.} \bibinfo{year}{2014}\natexlab{}.
\newblock \showarticletitle{Exploratory font selection using crowdsourced
  attributes}.
\newblock \bibinfo{journal}{\emph{ACM Trans. Graph.}} (\bibinfo{date}{July}
  \bibinfo{year}{2014}).
\newblock


\bibitem[\protect\citeauthoryear{Paszke, Gross, Chintala, Chanan, Yang, DeVito,
  Lin, Desmaison, Antiga, and Lerer}{Paszke et~al\mbox{.}}{2017}]%
        {paszke2017automatic}
\bibfield{author}{\bibinfo{person}{Adam Paszke}, \bibinfo{person}{Sam Gross},
  \bibinfo{person}{Soumith Chintala}, \bibinfo{person}{Gregory Chanan},
  \bibinfo{person}{Edward Yang}, \bibinfo{person}{Zachary DeVito},
  \bibinfo{person}{Zeming Lin}, \bibinfo{person}{Alban Desmaison},
  \bibinfo{person}{Luca Antiga}, {and} \bibinfo{person}{Adam Lerer}.}
  \bibinfo{year}{2017}\natexlab{}.
\newblock \showarticletitle{Automatic differentiation in PyTorch}. In
  \bibinfo{booktitle}{\emph{NIPS Workshop}}.
\newblock


\bibitem[\protect\citeauthoryear{Simonyan and Zisserman}{Simonyan and
  Zisserman}{2015}]%
        {simonyan_very_2015}
\bibfield{author}{\bibinfo{person}{Karen Simonyan} {and}
  \bibinfo{person}{Andrew Zisserman}.} \bibinfo{year}{2015}\natexlab{}.
\newblock \showarticletitle{Very {Deep} {Convolutional} {Networks} for
  {Large}-{Scale} {Image} {Recognition}}.
\newblock \bibinfo{journal}{\emph{arXiv:1409.1556 [cs]}} (\bibinfo{date}{April}
  \bibinfo{year}{2015}).
\newblock


\bibitem[\protect\citeauthoryear{Srivastava, Hinton, Krizhevsky, Sutskever, and
  Salakhutdinov}{Srivastava et~al\mbox{.}}{2014}]%
        {DBLP:journals/jmlr/SrivastavaHKSS14}
\bibfield{author}{\bibinfo{person}{Nitish Srivastava},
  \bibinfo{person}{Geoffrey~E. Hinton}, \bibinfo{person}{Alex Krizhevsky},
  \bibinfo{person}{Ilya Sutskever}, {and} \bibinfo{person}{Ruslan
  Salakhutdinov}.} \bibinfo{year}{2014}\natexlab{}.
\newblock \showarticletitle{Dropout: a simple way to prevent neural networks
  from overfitting}.
\newblock \bibinfo{journal}{\emph{J. Mach. Learn. Res.}} \bibinfo{volume}{15},
  \bibinfo{number}{1} (\bibinfo{year}{2014}), \bibinfo{pages}{1929--1958}.
\newblock
\urldef\tempurl%
\url{http://dl.acm.org/citation.cfm?id=2670313}
\showURL{%
\tempurl}


\bibitem[\protect\citeauthoryear{Szegedy, Liu, Jia, Sermanet, Reed, Anguelov,
  Erhan, Vanhoucke, and Rabinovich}{Szegedy et~al\mbox{.}}{2015}]%
        {Szegedy_2015_CVPR}
\bibfield{author}{\bibinfo{person}{Christian Szegedy}, \bibinfo{person}{Wei
  Liu}, \bibinfo{person}{Yangqing Jia}, \bibinfo{person}{Pierre Sermanet},
  \bibinfo{person}{Scott Reed}, \bibinfo{person}{Dragomir Anguelov},
  \bibinfo{person}{Dumitru Erhan}, \bibinfo{person}{Vincent Vanhoucke}, {and}
  \bibinfo{person}{Andrew Rabinovich}.} \bibinfo{year}{2015}\natexlab{}.
\newblock \showarticletitle{Going Deeper With Convolutions}. In
  \bibinfo{booktitle}{\emph{Proceedings of the IEEE Conference on Computer
  Vision and Pattern Recognition (CVPR)}}.
\newblock


\bibitem[\protect\citeauthoryear{Tao, Jin, Zhang, Yang, and Wang}{Tao
  et~al\mbox{.}}{2014}]%
        {tao_sparse_2014}
\bibfield{author}{\bibinfo{person}{Dapeng Tao}, \bibinfo{person}{Lianwen Jin},
  \bibinfo{person}{Shuye Zhang}, \bibinfo{person}{Zhao Yang}, {and}
  \bibinfo{person}{Yongfei Wang}.} \bibinfo{year}{2014}\natexlab{}.
\newblock \showarticletitle{Sparse {Discriminative} {Information}
  {Preservation} for {Chinese} character font categorization}.
\newblock \bibinfo{journal}{\emph{Neurocomputing}} (\bibinfo{date}{April}
  \bibinfo{year}{2014}).
\newblock


\bibitem[\protect\citeauthoryear{Wang, Yang, Jin, Shechtman, Agarwala, Brandt,
  and Huang}{Wang et~al\mbox{.}}{2015}]%
        {wang_deepfont_2015}
\bibfield{author}{\bibinfo{person}{Zhangyang Wang}, \bibinfo{person}{Jianchao
  Yang}, \bibinfo{person}{Hailin Jin}, \bibinfo{person}{Eli Shechtman},
  \bibinfo{person}{Aseem Agarwala}, \bibinfo{person}{Jonathan Brandt}, {and}
  \bibinfo{person}{Thomas~S. Huang}.} \bibinfo{year}{2015}\natexlab{}.
\newblock \showarticletitle{{DeepFont}: {Identify} {Your} {Font} from {An}
  {Image}}. In \bibinfo{booktitle}{\emph{Proceedings of the 23rd {ACM}
  international conference on {Multimedia}}}. \bibinfo{publisher}{Association
  for Computing Machinery}.
\newblock


\bibitem[\protect\citeauthoryear{Yang, Kim, Kwak, and Kim}{Yang
  et~al\mbox{.}}{2019}]%
        {DBLP:journals/ijdar/YangKKK19}
\bibfield{author}{\bibinfo{person}{Jinhyeok Yang}, \bibinfo{person}{Heebeom
  Kim}, \bibinfo{person}{Hyobin Kwak}, {and} \bibinfo{person}{Injung Kim}.}
  \bibinfo{year}{2019}\natexlab{}.
\newblock \showarticletitle{HanFont: large-scale adaptive Hangul font
  recognizer using {CNN} and font clustering}.
\newblock \bibinfo{journal}{\emph{Int. J. Document Anal. Recognit.}}
  \bibinfo{volume}{22}, \bibinfo{number}{4} (\bibinfo{year}{2019}),
  \bibinfo{pages}{407--416}.
\newblock
\urldef\tempurl%
\url{https://doi.org/10.1007/s10032-019-00337-w}
\showDOI{\tempurl}


\bibitem[\protect\citeauthoryear{Zhang, Guo, and Fan}{Zhang
  et~al\mbox{.}}{2019}]%
        {DBLP:journals/ijdar/ZhangGF19}
\bibfield{author}{\bibinfo{person}{Jiulong Zhang}, \bibinfo{person}{Mingtao
  Guo}, {and} \bibinfo{person}{Jianping Fan}.} \bibinfo{year}{2019}\natexlab{}.
\newblock \showarticletitle{A novel {CNN} structure for fine-grained
  classification of Chinese calligraphy styles}.
\newblock \bibinfo{journal}{\emph{Int. J. Document Anal. Recognit.}}
  \bibinfo{volume}{22}, \bibinfo{number}{2} (\bibinfo{year}{2019}),
  \bibinfo{pages}{177--188}.
\newblock
\urldef\tempurl%
\url{https://doi.org/10.1007/s10032-019-00324-1}
\showDOI{\tempurl}


\end{thebibliography}

\end{document}